\documentclass[letterpaper, 10 pt, journal, twoside]{IEEEtran}

\usepackage{graphics} 
\usepackage{epsfig} 
\usepackage{times} 
\usepackage{amsmath} 
\usepackage{amssymb}  
\usepackage{booktabs}
\usepackage{stfloats}
\usepackage[hidelinks]{hyperref}
\usepackage{float}
\usepackage{color}
\usepackage{etoolbox}
\makeatletter
\patchcmd{\@makecaption}
  {\scshape}
  {}
  {}
  {}
\makeatletter
\patchcmd{\@makecaption}
  {\\}
  {.\ }
  {}
  {}
\makeatother
\begin{document}
\title{Foldsformer: Learning Sequential Multi-Step Cloth Manipulation with Space-Time Attention}

\author{Kai Mo$^{1}$, Chongkun Xia$^{1}$, Xueqian Wang$^{1}$, Yuhong Deng$^{1}$, Xuehai Gao$^{2}$ and Bin Liang$^{3}$
\thanks{Manuscript received: September, 2, 2022; Revised November, 18, 2022; Accepted December, 9, 2022. 
This paper was recommended for publication by Editor Markus Vincze upon evaluation of the Associate Editor and Reviewers' comments.
This work was supported by the National Key R\&D Program of China (2022YFB4701400/4701402), National Natural Science Foundation of China (No. U1813216, U21B6002, 62203260), the Shenzhen Science Fund for Distinguished Young Scholars (RCJC20210706091946001), Guangdong Young Talent with Scientific and Technological Innovation (2019TQ05Z111), China Postdoctoral Science Foundation (2022M711823). (\textit{Corresponding author: Chongkun Xia \& Xueqian Wang.})} 
\thanks{$^{1}$Kai Mo, Chongkun Xia, Xueqian Wang, and Yuhong Deng are with the Center for Intelligent Control and Telescience, Tsinghua Shenzhen International Graduate School, 518055 Shenzhen, China. (email: mok21@mails.tsinghua.edu.cn, xiachongkun@sz.tsinghua.edu.cn, wang.xq@sz.tsinghua.edu.cn, dengyh20@mails.tsinghua.edu.cn)
}
\thanks{$^{2}$Xuehai Gao is with the Research Institute of Tsinghua University in Shenzhen, 518055 Shenzhen, China. (email: gaoxh@tsinghua-sz.org)}
\thanks{$^{3}$Bin Liang is with the Navigation and Control Research Center, Department of Automation, Tsinghua University, 100084 Beijing, China. (email: bliang@tsinghua.edu.cn)}
\thanks{Digital Object Identifier (DOI): see top of this page.}
}

\markboth{IEEE Robotics and Automation Letters. Preprint Version. Accepted December, 2022}
{Mo \MakeLowercase{\textit{et al.}}: Foldsformer: Learning Sequential Multi-Step Cloth Manipulation with Space-Time Attention}

\maketitle

\begin{abstract}
Sequential multi-step cloth manipulation is a challenging problem in robotic manipulation, requiring a robot to perceive the cloth state and plan a sequence of chained actions leading to the desired state. 
Most previous works address this problem in a goal-conditioned way, and goal observation must be given for each specific task and cloth configuration, which is not practical and efficient. Thus, we present a novel multi-step cloth manipulation planning framework named Foldformer. Foldformer can complete similar tasks with only a general demonstration and utilize a space-time attention mechanism to capture the instruction information behind this demonstration.
We experimentally evaluate Foldsformer on four representative sequential multi-step manipulation tasks and show that Foldsformer significantly outperforms state-of-the-art approaches in simulation. Foldformer can complete multi-step cloth manipulation tasks even when configurations of the cloth (e.g., size and pose) vary from configurations in the general demonstrations. Furthermore, our approach can be transferred from simulation to the real world without additional training or domain randomization. Despite training on rectangular clothes, we also show that our approach can generalize to unseen cloth shapes (T-shirts and shorts). Videos and source code are available at: \url{https://sites.google.com/view/foldsformer}.
\end{abstract}

\begin{IEEEkeywords}
Deep Learning in Grasping and Manipulation, Perception-Action Coupling
\end{IEEEkeywords}

\section{INTRODUCTION}
\IEEEPARstart{C}{loth} manipulation has wide applications in domestic, medical and industrial settings, such as laundry-folding~\cite{FoldingPipline,laundryfolding}, surgery~\cite{surgery} and manufacturing~\cite{industrial}. Most of these tasks can be seen as sequential multi-step cloth manipulation, where sequential chained actions should be planned to achieve the desired cloth state. There are some challenges for sequential multi-step cloth manipulation: Unlike rigid objects, cloth has infinite degrees of freedom, making it hard for state representation; The cloth dynamics is also complex~\cite{nonlinear}, and slightly different interactions may lead to significantly different cloth behaviors; Further, a single counter-productive action may crumple the cloth and make it difficult to recover~\cite{rearrange}, which means completing a sequential multi-step cloth manipulation task requires only a particular set of chained actions.

\begin{figure}[!t]
    \centering
    \includegraphics[width=0.475\textwidth]{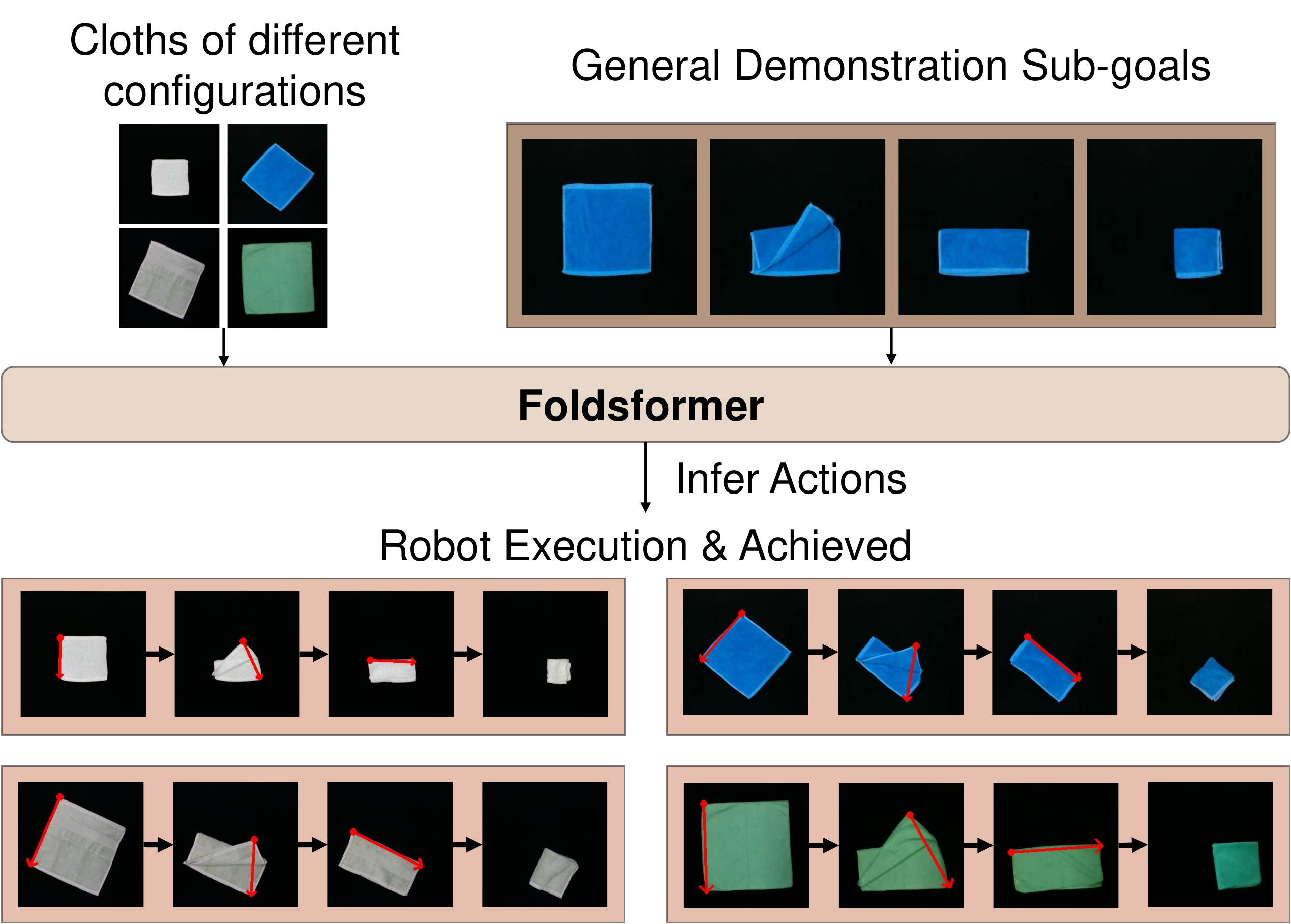}
    \setlength{\abovecaptionskip}{-0.1cm}
    \caption{\textbf{Overview.} Foldsformer can complete similar tasks even when the cloth configurations vary from that in a genral demonstration.} 
    \label{intro}
    \vspace{-0.5cm}
\end{figure}

For multi-step cloth manipulation, early works designed hard-coded policies with task-customized grippers and cloth with markers, making it hard to generalize these policies to other settings~\cite{FoldingPipline,laundryfolding}. Recently, learning-based approaches have shown great potential for cloth manipulation by tackling the difficulties in state estimation or modeling with neural networks. Some of these approaches learn task-specific sequential multi-step cloth manipulation policies from expert demonstrations. Although these approaches are demonstrated to be effective, different models should be trained to perform different tasks~\cite{deepIL,boxwrapping,dmfd}. Others learn a goal-conditioned policy from random data and can achieve arbitrary cloth goal states. Lee~\textit{et al.}~\cite{lee} specifies the goal state with a single image at test time. However, for a sequential multi-step cloth manipulation task, the cloth of the goal state can be highly self-occluded, and a single image fails to contain enough information about the goal. Weng~\textit{et al.}~\cite{fabricflownet} specifies the goal state with a sequence of sub-goal images from an expert demonstration at test time.
For the goal-conditioned policies mentioned above, the goal images must be given for each specific task and cloth configuration by human demonstrators, which is not practical and efficient. When configurations of the cloth vary from that in the goal, these policies' performances greatly downgrade.

In this work, we address the problem mentioned above in sequential multi-step cloth manipulation. We propose a novel framework named Foldsformer that can complete similar tasks of different cloth configurations with a general demonstration.
We design a novel encoder in Foldsformer based on the space-time attention mechanism. Given a general demonstration containing a sequence of sub-goals that describe a multi-step task, this encoder can capture the instruction information behind this demonstration that helps Foldsfomer to perform similar tasks even when the cloth configuration varies (see Fig.~\ref{intro}).
We first leverage large-scale pre-training from random actions and then transfer it to different tasks through only 100 expert demonstrations per task fine-tuning. 
Experimental results show that our approach greatly outperforms three previous state-of-the-art baselines~\cite{lee,fabricflownet,fabric_vsf}. We then demonstrate that our approach trained in simulation can be transferred to the real world without additional training or domain randomization. We also show that our learned model can successfully generalize to T-shirts and shorts despite being trained on rectangular clothes. In summary:
\begin{itemize}
    \item We propose a novel framework named Foldsformer for sequential multi-step manipulation. Foldsformer can complete similar tasks on clothes of different configurations with a general demonstration.
    
    \item We propose a novel encoder architecture in Foldsformer based on space-time attention. This encoder enables a robot to learn instruction information behind a general demonstration.
    
    \item We conduct experiments to demonstrate that Foldsformer can be zero-shot transferred to the real world and generalize to unseen T-shirts and shorts despite being trained only on rectangular cloths.
\end{itemize}

\section{Related Work}
\subsection{Sequential Multi-Step Cloth Manipulation}
There are two mainstream approaches for sequential multi-step cloth manipulation: analytical approaches and learning-based approaches. Analytical approaches are mainly task-specific, these approaches manually divide a task into several steps and then design scripted actions for each step. Doumanoglou~\textit{et al.}~\cite{FoldingPipline} and Bersch~\textit{et al.}~\cite{laundryfolding} design complete pipelines for cloth folding. Yamakawa~\textit{et al.}~\cite{dynamicscripted} achieve dynamic folding of cloth with custom high-speed robot hands and high-speed sliders. Miller~\textit{et al.}~\cite{geometricfolding} model cloth as as polygons and complete folding tasks with a model-based optimization approach. However, these analytical approaches mainly use task-customized robot systems to adapt to their scripted actions. It is hard to generalize them to other settings.

Learning-based approaches can be divided into two categories: model-based and model-free. Model-based approaches learn a forward cloth dynamic model and use it for planning~\cite{cfm,fabric_vsf,graphdynamics,vcd}. However, these model-based approaches require time-consuming planning like Cross Entropy Method or Model Predictive Control to achieve the goal cloth state. As for model-free approaches, some approaches learn task-specific cloth manipulation policies from expert demonstrations~\cite{deepIL,boxwrapping,dmfd}. However, these task-specific policies fail to reuse information for different tasks efficiently. Some approaches learn a goal-conditioned policy that can achieve arbitrary unseen cloth states purely from random data. Lee~\textit{et al.}~\cite{lee} learn a goal-conditioned pick-and-place policy with standard Deep Q-network and use predefined sets of fold angles and fold distances to ensure learning efficiency, but this discrete action space limits its manipulation ability.
Weng~\textit{et al.}~\cite{fabricflownet} propose a flow-based policy and divide a multi-step goal into a sequence of sub-goals at test time, making it easier to perform multi-step tasks. However, when the cloth configuration varies from that in demonstration sub-goal images, the flow estimation can be disturbed by the variation of the cloth configurations. Hence, the flow-based representation fails to describe the change in the cloth states between the current observation and the sub-goal.
Ganapathi~\textit{et al.}~\cite{densedescriptor} learns dense visual correspondences between cloths of different size, color, and pose for sequential multi-step cloth manipulation. However, it relies on demonstration observations and actions provided by a human expert at test time. 
In contrast, our approach only needs general demonstration observations and can perform similar tasks on different cloth configurations without accessing the expert actions.

\subsection{Space-time Attention}
The large-scale adoption of the attention mechanism was first proposed by Transformer architecture~\cite{transformer} in natural language processing and achieved great success. The attention mechanism can capture both local and global contexts with less domain-specific inductive bias, making it easily applied to some other domains. Vision Transformer~\cite{vit} extends Transformer architecture to image classification tasks by splitting images into sequences of image patches. Bertasius~\textit{et al.}~\cite{timesformer} propose TimeSformer and adapt the standard Transformer to video with space-time attention mechanism. The space-time attention mechanism extracts a video's temporal and spatial features separately with a great speed-accuracy trade-off. By leveraging space-time attention, TimeSformer is also faster to train than traditional CNN video architecture and achieves higher efficiency.
By utilizing the space-time attention mechanism, we extract the spatiotemporal features of a sequence of subgoals to capture the instruction information behind the demonstration.

\section{Approach}
\subsection{Problem Formulation}
Our goal is to enable a robot to perform sequential multi-step cloth manipulation tasks under the instruction of a general demonstration. Let each task be defined by a sequence of sub-goal observations $\mathcal{G} : \{x_1^g, x_2^g, ..., x_N^g\}$, these sub-goal observations are obtained from an expert demonstration. Rather than a single goal, we use sub-goals because the cloth of the goal state in sequential multi-step cloth manipulation tasks can be highly self-occluded. A single goal loses part of the information about the full goal state. Unlike prior works~\cite{fabricflownet,rope,zeroshot}, we do not require the cloth configuration in sub-goals to be the same as the cloth configuration at test time. A general demonstration should be applied in different settings.

We formulate the problem as learning a policy $\pi$ that infers a pick-and-place action $a_t \in \mathcal{A}$ from the current visual observation of the cloth $o_t \in \mathcal{O}$ and the sub-goals $\mathcal{G}$:
\begin{equation}
    \pi(o_t, \mathcal{G}) \rightarrow a_t = (p_{\rm pick},p_{\rm place}) \in \mathcal{A}
\end{equation}
where $p_{\rm pick}$ is the pixel pick point of the end effector when grasping part of the cloth, and $p_{\rm place}$ is the pixel place point of the end effector when releasing the grasp. 

\subsection{Foldsformer}
\label{approach}

\begin{figure*}[t]
    \centering
    \includegraphics[width=0.9\textwidth]{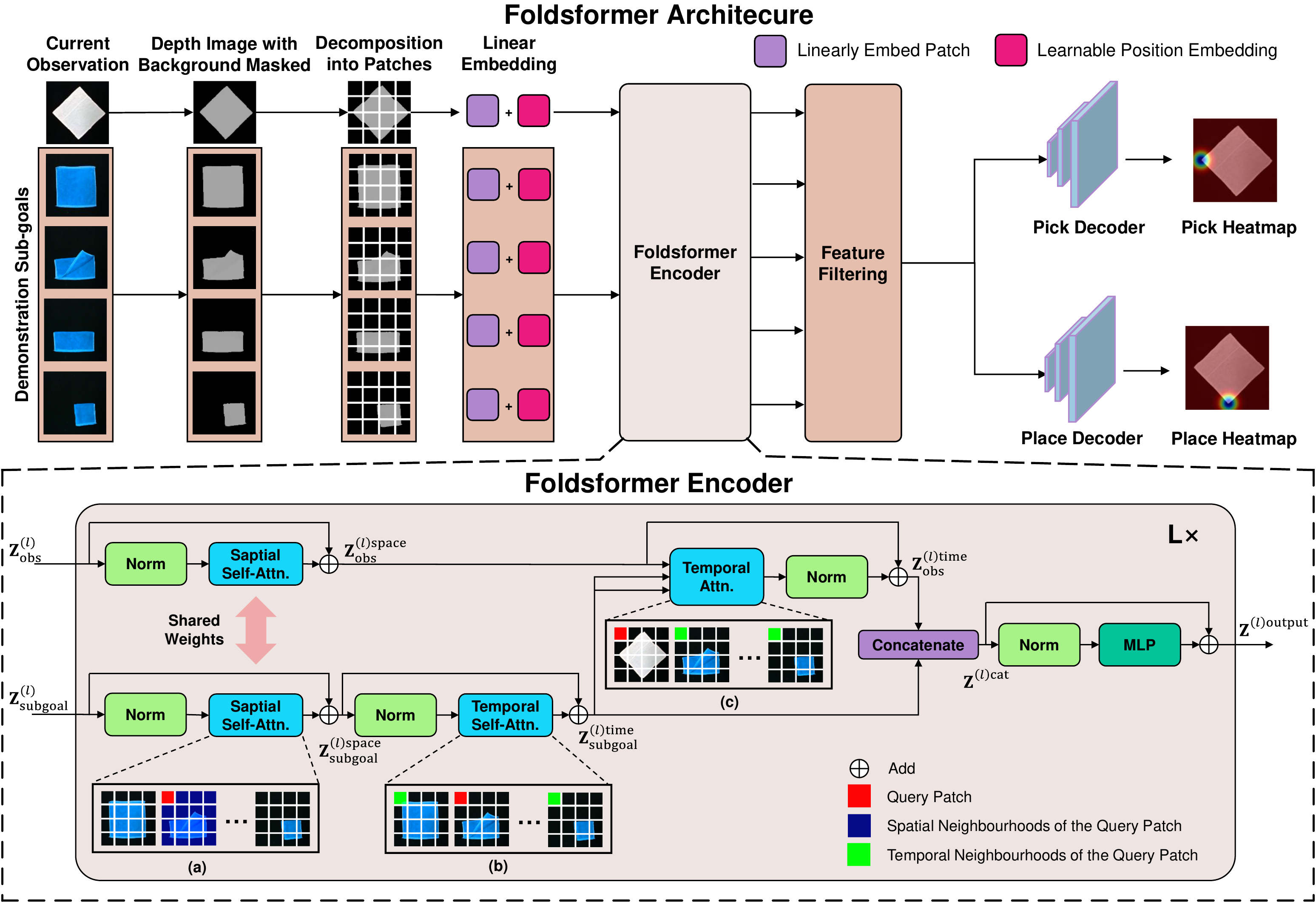}
    \caption{\textbf{Foldsformer Architecture.} Foldsformer takes the current observation depth image and the sub-goal depth image sequence as input and outputs a pick heatmap and a place heatmap that simultaneously estimates a pick point and a place point. In the Foldsformer encoder block, we split the input into two vectors and process them in different ways, and we finally make a fusion to aggregate information. For illustration, we denote the query patch in red, its spatial neighborhoods in blue, and its temporal neighborhoods in green. Patches without color are not used for the attention computation of the red patch. (a) Visualization of the space multi-head self-attention. The self-attention is computed for every patch in all images. (b) Visualization of the time multi-head self-attention. The self-attention is computed for every patch in sub-goal images. (c) Visualization of the time multi-head attention. The self-attention is computed for every patch in the current observation image.}
    \label{foldformer}
    \vspace{-0.5cm}
\end{figure*}

Previous approaches input the current observation $o_t$ and each sub-goal $x_i^g$ sequentially to their goal-conditioned policies~\cite{fabricflownet,rope,zeroshot}. In this way, these policies can not access a complete picture of the multi-step task, resulting in the loss of the temporal information of the sub-goal sequence. Hence, these policies can only capture limited instruction information from the demonstration sub-goals, as they can only access each sub-goal separately. 
Our insight is that we can allow the policy to access all the sub-goals together and take advantage of the temporal features of all sub-goal images to obtain more instruction information about a multi-step task.

The schematic of the proposed Foldsformer architecture can be found in Fig.~\ref{foldformer}. Foldsformer is a variant of TimeSformer~\cite{timesformer} in sequential multi-step cloth manipulation. Foldsformer takes the current observation $o_t$ and the demonstration sub-goal observation sequence $\mathcal{G}$ as input and outputs a pick heatmap and a place heatmap estimating a pick point and a place point simultaneously. Further details of Foldsformer are described below.

\textbf{Foldsformer input.} Foldsformer takes as input a sequence of depth images $X = \{o_t, \mathcal{G}\}\in \mathbb{R}^{H \times W \times 1 \times (F+1)} $ that consists of the current observation depth image $o_t$ of size $H \times W$ and $F$ sub-goal depth images $\mathcal{G}$ of size $H \times W$. We use depth rather than RGB inspired by~\cite{fabricflownet}, because the gap between simulated depth images and real depth images is small. By using depth-only input, our approach can be easily transferred to the real world without any domain randomization.

\textbf{Decomposition into patches.} We decompose each image into $N$ non-overlapping patches, each of size $P \times P$, then we flatten these patches into vectors ${\rm \textbf{x}}_{(p,t)} \in \mathbb{R}^{P^2}$ with $p = 1, ..., N$ denoting spatial locations and $t = 1, ..., F+1$ denoting an index over images.

\textbf{Linear Embedding.} We map each patch ${\rm \textbf{x}}_{(p,t)}$ into an embedding vector ${\rm \textbf{z}}_{(p,t)} \in \mathbb{R}^D$ of $D$ dimensions linearly:
\begin{equation}
   {\rm \textbf{z}}^{(0)}_{(p,t)} = E{\rm \textbf{x}}_{(p,t)} + {\rm \textbf{e}}^{pos}_{(p,t)}
\end{equation}
where $E \in \mathbb{R}^{D \times P^2}$ is a learnable matrix, ${\rm \textbf{e}}^{pos}_{(p,t)} \in \mathbb{R}^D$ denotes a learnable position embedding that encodes the spatial and temporal position of each patch.

\textbf{Foldsformer Encoder.} We design our Foldsformer encoder block (see Fig.~\ref{foldformer}) based on the divided space-time attention mechanism in TimeSformer~\cite{timesformer}. Compared to the TimeSformer encoder block, Foldsformer encoder block first splits the input into vectors corresponding to the current observation and vectors corresponding to the demonstration sub-goal sequence, then processes them in different ways and finally makes a fusion of them to aggregate information. We process them separately because the cloth's configuration in the current observation may differ from that in the demonstration sub-goal sequence.

Let ${\textbf{z}}^{(l)\rm{input}} \in \mathbb{R}^{\frac{H}{P} \times \frac{W}{P} \times D \times (F+1)}$ denote the input vectors of the encoder block $l$, with $l = 1, ..., L$ denoting an index over blocks. ${\textbf{z}}^{(l)\rm{input}}$ can be split in time dimension into vectors ${\textbf{z}}_{\rm obs}^{(l)} \in \mathbb{R}^{\frac{H}{P} \times \frac{W}{P} \times D \times 1}$ that correspond to the current observation image and ${ \textbf{z}}_{\rm subgoal}^{(l)} \in \mathbb{R}^{\frac{H}{P} \times \frac{W}{P} \times D \times F}$ that correspond to the sub-goal image sequence.

We firstly compute multi-head self-attention~\cite{vit} of vectors ${\textbf{z}}_{\rm obs}^{(l)}$ and ${\textbf{z}}_{\rm subgoal}^{(l)}$ with the same weights in space dimension. Each patch and its spatial neighbors within the same image are used for the space self-attention computation (see Fig.~\ref{foldformer}(a)).
We use residual connection after space self-attention to aggregate information:
\begin{equation}
\begin{split}
    \rm {\textbf{z}}_{obs}^{(\textit{l})space} &= \rm SpaceMSA(LN(\textbf{z}_{obs}^{(\textit{l})})) + \textbf{z}_{obs}^{(\textit{l})} 
    \\
    \rm \textbf{z}_{subgoal}^{(\textit{l})space} &= \rm SpaceMSA(LN(\textbf{z}_{subgoal}^{(\textit{l})})) + \textbf{z}_{subgoal}^{(\textit{l})}
\end{split}
\end{equation}
where $\rm LN()$ denotes LayerNorm~\cite{layernorm}, and $\rm SpaceMSA()$ denotes space multi-head self-attention. This space multi-head self-attention extracts spatial information corresponding to the cloth state in the current observation image and each sub-goal image.

We then compute multi-head self-attention of vectors ${\rm \textbf{z}_{subgoal}^{(\textit{l})space}}$ in time dimension. Each patch and its temporal neighbors at the same spatial location are used for the time self-attention computation (see Fig. ~\ref{foldformer}(b)). We use residual connection after time self-attention to aggregate information:
\begin{equation}
    \rm \textbf{z}_{subgoal}^{(\textit{l})time} = TimeMSA(LN(\textbf{z}_{subgoal}^{(\textit{l})space})) + \textbf{z}_{subgoal}^{(\textit{l})space}
\end{equation}
where $\rm TimeMSA()$ denotes time multi-head self-attention. This time multi-head self-attention extracts temporal information corresponding to the cloth state changes among all sub-goal images.

After space multi-head self-attention and time multi-head self-attention, we then add a time multi-head attention (see Fig.~\ref{foldformer}(c)) to fuse the extracted information of the current observation and the sub-goals. The time multi-head attention is computed in time dimension (see Fig.~\ref{foldformer}(c)), where $\rm \textbf{z}_{obs}^{(\textit{l})space}$ serves as the query, $\rm \textbf{z}_{subgoal}^{(\textit{l})time}$ serves as the key and the value. We use residual connection after the time multi-head attention to aggregate information:
\begin{equation}
\begin{aligned}
    \rm \textbf{z}_{obs}^{(\textit{l})time} &= \rm LN(TimeMA(\textbf{z}_{obs}^{(\textit{l})space},\textbf{z}_{subgoal}^{(\textit{l})time},
    \\ & \rm \textbf{z}_{subgoal}^{(\textit{l})time}))  
     + \rm \textbf{z}_{obs}^{(\textit{l})space}
\end{aligned}
\end{equation}
where $\rm TimeMA()$ denotes time multi-head attention.

Then, the concatenation $\textbf{z}^{(l)\rm cat}$ of  $\rm \textbf{z}_{obs}^{(\textit{l})time}$ and $\rm \textbf{z}_{subgoal}^{(\textit{l})time}$ is projected and passed through an MLP, using residual connection:
\begin{equation}
\begin{aligned}
    \textbf{z}^{(l)\rm cat} &= \rm Concatenate(\textbf{z}_{obs}^{(\textit{l})time},\textbf{z}_{subgoal}^{(\textit{l})time}) 
    \\
    \textbf{z}^{(l)\rm output} &= \rm MLP(LN( \textbf{z}^{(l)cat})) + \textbf{z}^{(l)cat}
\end{aligned}
\end{equation}

The output of the Foldsformer encoder block $\textbf{z}^{(l)\rm output}$ keeps the same dimension as the input of the Foldsformer encoder block $\textbf{z}^{(l)\rm input}$, so these encoder blocks can be stacked in a sequence to make the network deep. Our encoder consists of several Foldsformer encoder blocks.

\textbf{Feature filtering.} In the Foldsformer encoder output, we only use the vectors that correspond to the current observation and vectors that correspond to the last image in the sub-goal image sequence, i.e., the goal image. Let ${\rm \textbf{z}}^{\rm output}_{(t)} \in \mathbb{R}^{\frac{H}{P} \times \frac{W}{P} \times D \times (F+1)}$ denote the encoder output, where $t = 1, ..., F+1$ denotes an index over images. We choose ${\rm \textbf{z}}^{\rm output}_{(1)}$ and ${\rm \textbf{z}}^{\rm output}_{(F+1)}$ and concatenate them for the decoder input:
\begin{equation}
    \rm \textbf{z}^{filter} = Concatenate(\textbf{z}^{output}_{(1)},\textbf{z}^{output}_{(\textit{F}+1)})
\end{equation}
In this way, the decoder only takes as input the feature of the current observation and the goal image to infer an action, forming closed loop feedback.

\textbf{Decoder.} We build a pick decoder and a place decoder to estimate a pick and place point, respectively. The two decoders have the same architecture. Similar to~\cite{pup}, we consider a progressive upsampling strategy that alternates convolutional layers and upsampling operations. The two decoders take $\rm \textbf{z}^{filter}$ as input, then output a predicted pick heatmap $Q_{\rm pick}$ and a predicted place heatmap $Q_{\rm place}$ that have the same size as the input images. From the two heatmaps, we can get the optimal pick point and the optimal place point:
\begin{equation}
\begin{split}
    p_{\rm pick} &= {\rm argmax}_{p} Q_{\rm pick}(p)
   \\
    p_{\rm place} &= {\rm argmax}_{p} Q_{\rm place}(p)
\end{split}
\end{equation}

\textbf{Loss.} We supervise the whole network using the binary cross-entropy(BCE) loss between predicted heatmaps $Q_{\rm pick}$, $Q_{\rm place}$ and groud truth heatmaps $Q_{\rm pick}^{\rm gt}$, $Q_{\rm place}^{\rm gt}$, and compute the sum of them:
\begin{equation}
 \rm   \mathcal{L} = BCE(\textit{Q}_{pick},\textit{Q}_{pick}^{gt}) + BCE(\textit{Q}_{place},\textit{Q}_{place}^{gt})
\end{equation}

\subsection{Implementation Details}
\label{simulator}
We collect training data and evaluate Foldsformer and the baselines in the SoftGym suite~\cite{softgym}, a particle-based simulator built on Nvidia Flex. In our simulation environment, a top-down camera provides depth and RGB images of size $224 \times 224$, and the cloth is put at the center of the camera view.

Our training data consists of samples generated by random actions and expert demonstrations. For collecting random training samples, we first generate 1000 random initial configurations of rectangular cloths whose size and pose are randomized. Then we collect 6000 trajectories of length 8 using random actions. Inspired by~\cite{fabricflownet}, 80\% of the random actions are biased to pick corners, and 20\% of the pick actions are uniformly sampled over the cloth. We use corner bias because most sequential multi-step cloth manipulation tasks tend to pick corners. For collecting expert demonstrations, we use scripted demonstrator policies to get 100 trajectories of expert demonstrations per task. In SoftGym, the cloth is modeled as a grid of particles, and the ground truth position of each particle can be accessed. Thus, we can easily get the true corners and script a demonstrator.

In each trajectory, we take a sequence of actions and record the initial observation $o_0$, actions $a_i = (p_{\rm pick}^{(i)},p_{\rm place}^{(i)})$, and the next observation $o_i$ after action $a_i$, with $i = 1,2,...,M$ denoting an index over actions. Mathematically, a trajectory $\mathcal{T}$ of length $M$ consists of a sequence of observations and actions:
\begin{equation}
    \mathcal{T} = \{o_0, a_1, o_1, ..., o_{M-1},a_M,o_M\}
\end{equation}

To train Foldsformer faster, we split each trajectory whose length is $M$ into sub-trajectories whose length is $K$. Let $\mathcal{S}_i = \{o_i,a_{i+1},o_{i+1},...,o_{i+K-1},a_{i+K},o_{i+K} | i=0,1,..,M-K\}$ denote a sub-trajectory. From a sub-trajectory, we can get $K$ training samples: the input $(o_{i+j},o_i,o_{i+1},...,o_{i+K})$ and the action $a_{i+j+1}$ used to supervise the network with $j = 0,1,...,K-1$. In this way, we can get $(M-K+1)H$ training samples from each trajectory of length $M$, making our approach sample-efficient. 

In this work, the patch size ($P$), the embedding vectors' dimension ($D$), the number of encoder blocks ($L$), and the length of the sub-trajectory ($K$) are set to 16, 256, 8, 4, respectively. Each encoder alternates 5 convolutional layers and 4 bilinear upsampling layers. The kernel sizes, strides, and filter sizes of the convolutional layers are [1, 1, 1, 1, 1], [1, 1, 1, 1, 1], [256, 256, 128, 128, 1], respectively. The scale factors of the bilinear upsampling layers are set to 2. We train Foldsformer with a batch size of 32, a learning rate of 0.0001, and an Adam optimizer~\cite{adam} for 20 epochs. It takes about 9 hours on an NVIDIA RTX 3090 GPU, and the BCE loss drops to about 0.02.

\begin{table*}[t]
\centering
\caption{Mean particle distance error (mm) and inference time (sec) on four sequential multi-step cloth manipulation tasks in simulation. The best results are bolded. We bold the results of Foldsformer (Full Approach) and Foldsformer (NoTimeAttn) in the \textit{Double Straight} task that have a significant overlap.}
\begin{tabular}{@{}lcccc|c@{}}
\toprule
Approach          & \textit{Double Triangle}      & \textit{Double Straight}       & \textit{All Corners Inward}  & \textit{Corners Egdes Inward} & Inference Time \\ \midrule
FabricFlowNet~\cite{fabricflownet} & 102.20 $\pm$ 19.39 & 85.34 $\pm$ 19.67  & 33.64 $\pm$ 11.76 & 36.95 $\pm$ 9.95 & \textbf{$\sim$ 0.002}  \\
Lee \textit{et al.}~\cite{lee}  & 109.82 $\pm$ 39.96 & 114.71 $\pm$ 26.06 & 27.21 $\pm$ 11.47 & 70.67 $\pm$ 22.24 & $\sim$ 0.061 \\
Fabric-VSF~\cite{fabric_vsf} & 114.62 $\pm$ 35.46 & 116.94 $\pm$ 24.52 & 46.05 $\pm$ 23.38 & 51.82 $\pm$ 16.65 & $\sim$ 6.512 \\
Foldsformer (Goal-Conditioned) & 93.30 $\pm$ 43.35 & 80.81 $\pm$ 38.39 & 49.46 $\pm$ 19.55 & 56.66 $\pm$ 43.25 & $\sim$ 0.007\\
Foldsformer (NoTimeAttn) & 59.86 $\pm$ 44.58 & \textbf{60.90 $\pm$ 47.52} & 9.11 $\pm$ 3.09 & 22.97 $\pm$ 20.42 & $\sim$ 0.005 \\
\textbf{Foldsformer (Full Approach)} & \textbf{19.64 $\pm$ 17.08} & \textbf{59.09 $\pm$ 44.82} & \textbf{3.06 $\pm$ 1.83} & \textbf{8.11 $\pm$ 7.96} &  $\sim$ 0.011 \\ \bottomrule
\end{tabular}
\label{simulation results}
\vspace{-0.5cm}
\end{table*}

\section{Experiments}
\subsection{Simulation Setup}
\label{simulation setup}
As mentioned in Sec.~\ref{simulator},  we use SoftGym~\cite{softgym} simulator for training and evaluation. We evaluate Foldsformer and the baselines on 4 sequential multi-step cloth manipulation tasks, which can be achieved by 2, 3, 4, and 4 actions, respectively (see Fig.~\ref{simulation figure} for details.): (i) \textit{Double Triangle}; (ii) \textit{Double Straight}; (iii) \textit{All Corners Inward}; (iv) \textit{Corners Edges Inward}. We consider 40 different cloth configurations per task, combining 10 different sizes and 4 different rotation angles. For task (i), task (ii), and task (iv), we use square cloth, the 10 different sizes range from $31.25\rm{cm} \times 31.25 \rm{cm}$ to $36.875 \rm{cm} \times 36.875 \rm{cm}$ divided into $0.625\rm{cm} \times 0.625\rm{cm}$ intervals. For task (ii), we use rectangular cloth, and the 10 different sizes range from $31.25\rm{cm} \times 27.5\rm{cm}$ to $36.875\rm{cm} \times 32.5\rm{cm}$ divided into $0.625\rm{cm} \times 0.5\rm{cm}$ intervals. The 4 different rotation angles are $0^{\circ}$, $30^{\circ}$, $45^{\circ}$ and $60^{\circ}$. Given a sequence of demonstration sub-goal images where the square cloth size is $34.375\rm{cm} \times 34.375\rm{cm}$, the rectangle cloth size is $34.375\rm{cm} \times 27.5\rm{cm}$ and the rotation angle is $0^{\circ}$, a policy needs to perform on all 40 different cloth configurations. Our error metric is the mean particle position error between the cloth states achieved by the policy and a scripted demonstrator mentioned in Sec.~\ref{simulator} that can access the ground truth particle positions.

We compare our approach Foldsformer to three previous state-of-the-art baselines for sequential multi-step cloth manipulation: FabricFlowNet~\cite{fabricflownet}, which learns a flow-based policy; Lee~\textit{et al.}~\cite{lee}, which uses offline RL for arbitrary-goal cloth folding; Fabric-VSF~\cite{fabric_vsf}, which learns a forward cloth dynamic model in pixel space and use it for planning. We adapt FabricFlowNet to a single-arm variant. Although the original approaches~\cite{lee,fabric_vsf} use only a single goal image, we provide sub-goals for them.

To further evaluate the importance of the components of our approach, we conduct two ablations in simulation: A standard goal-conditioned variant of our approach (``Goal-Conditioned"), which only takes as input the current observation and each sub-goal separately; An variant of our approach (``NoTimeAttn"), which removes time attention in the Foldsformer encoder block described in Sec.~\ref{approach}.

For each task, we fine-tune our approach, the baselines, and the ablations with 100 expert demonstrations.

\subsection{Simulation Results}
\begin{figure*}[t]
    \centering
    \includegraphics[width=0.975\textwidth]{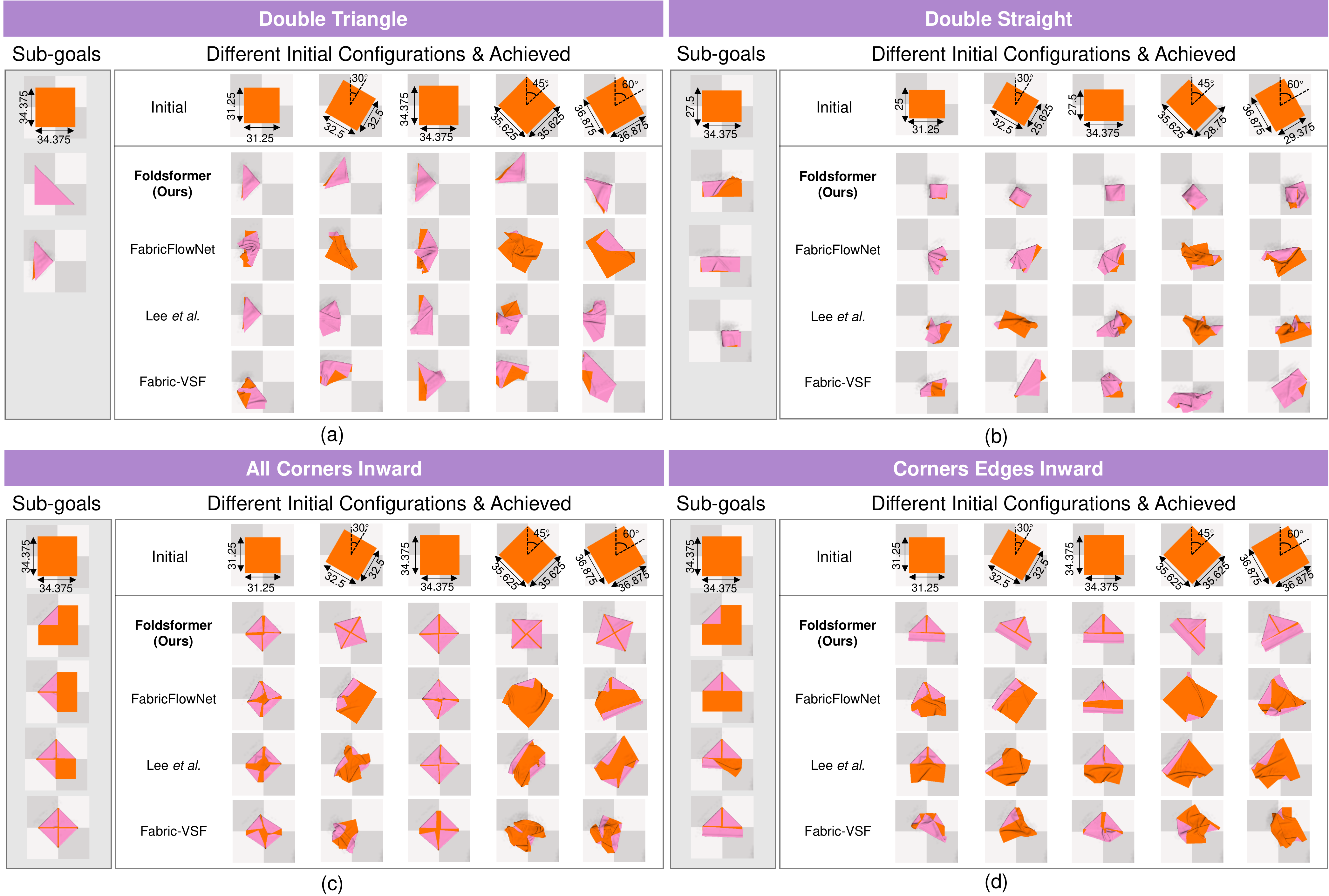}
    \setlength{\abovecaptionskip}{-0.1cm}
    \caption{\textbf{Qualitative Results in Simulation Experiments.} We evaluate Foldsformer and the baselines on 4 sequential multi-step cloth manipulation tasks: (a) \textit{Double Triangle.} (b) \textit{Double Straight.} (c) \textit{All Corners Inward.} (d) \textit{Corners Edges Inward.}}
    \label{simulation figure}
    \vspace{-0.5cm}
\end{figure*}

Simulation results can be found in Table.~\ref{simulation results}. Foldsformer achieves the lowest mean particle error over all tasks and shows a larger improvement over the baselines. Fig.~\ref{simulation figure} shows representative qualitative results on 5 of all 40 different clothes. As can be seen, even when the size and pose of the cloth are different from that of the cloth in demonstration sub-goals, Foldsformer performs all tasks successfully, while the baselines fail to complete most of the tasks. 
Although the inference time\footnote{The inference time is tested on one machine with Intel i9-9900K CPU (3.60GHz) and NVIDIA GeForce RTX 2080Ti} of Foldsformer is slower than FabicFlowNet ($\sim$0.002s vs. $\sim$0.011s), it is still acceptable in real-time applications.

For the ablations, the ``Goal-Conditioned" ablation validates the significance of allowing a policy to access all the sub-goals together. A policy can capture more instruction information by accessing all the sub-goals together rather than accessing them separately. The ``NoTimeAttn" ablation demonstrates the benefits of time attention to leverage the temporal features of the sub-goal images. Previous approaches input the sub-goals separately, leading to the loss of these temporal features.

\subsection{Real World Setup}
\begin{figure}[!b]
    \vspace{-0.5cm}
    \centering
    \includegraphics[width=0.475\textwidth]{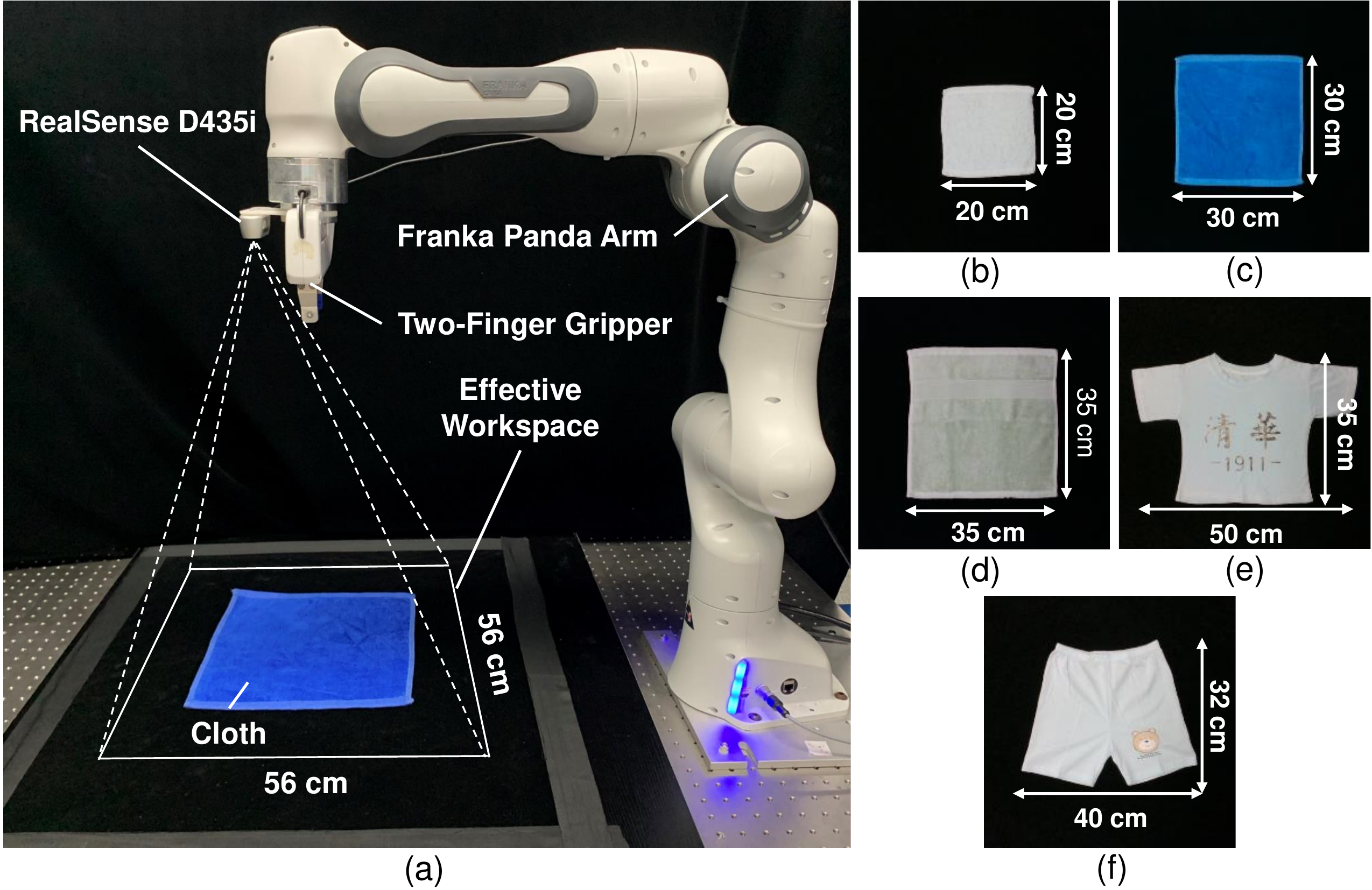}
    \caption{\textbf{Experimental Setup.} (a) Robot system. (b) A $20\rm{cm} \times 20\rm{cm}$ cloth. (c) A $30\rm{cm} \times 30\rm{cm}$ cloth.  (d) A $35\rm{cm} \times 35\rm{cm}$ cloth. (e) A $50\rm{cm} \times 35\rm{cm}$ T-shirt. (f) A pair of shorts ($40\rm{cm} \times 32\rm{cm}$).}
    \label{expermental setup}
\end{figure}

Fig.~\ref{expermental setup}(a) shows the robot system in the real world. We use a 7-DOF Franka Emika Panda robot arm with a standard two-finger panda gripper. The pick and place trajectories are planned by using MoveIt!~\cite{moveit}. We obtain RGB-D images from an Intel Realsense D435i camera mounted at the end of the arm. Before capturing images, the arm moves to a fixed pose where the camera can capture an effective workspace of $56\rm{cm} \times 56\rm{cm}$. To narrow the gap between simulated depth images and real-world depth images, we crop the center of the raw depth images from the camera and segment the cloth by color-thresholding the background. Similar to~\cite{fabricflownet}, we also add the difference between the simulated surface and the mean depth of the support surface in the real world to align the depth between real and simulated depth images. Through these simple techniques, our approach trained in simulation can be transferred to the real world successfully without additional training or domain randomization.

\begin{figure*}[t]
    \centering
    \includegraphics[width=1.0\textwidth]{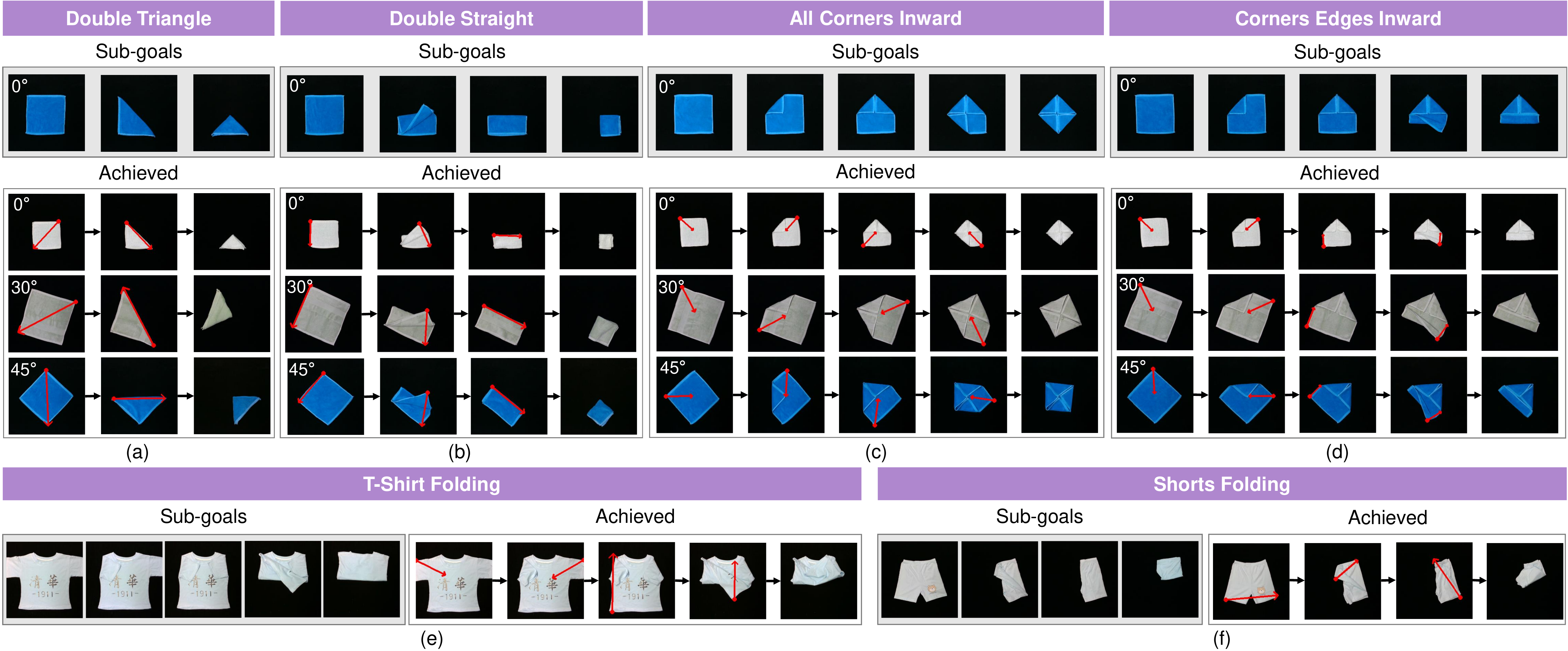}
    \setlength{\abovecaptionskip}{-0.5cm}
    \caption{\textbf{Qualitative Results in Real World Experiments.} Foldsformer performs 4 sequential multi-step cloth manipulation tasks on square cloths in the real world: (a) \textit{Double Triangle.} (b) \textit{Double Straight.} (c) \textit{All Corners Inward.} (d) \textit{Corners Edges Inward.} Foldsformer also completes a T-shirt folding task and a shorts folding task where the T-shirt and shorts are unseen cloth shapes during training: (e) \textit{T-shirt Folding.} (f) \textit{Shorts Folding.} Pick and place actions are visualized as red arrows.}
    \label{real world figure}
    \vspace{-0.5cm}
\end{figure*}

We use 3 square cloths of different size: a $20\rm{cm} \times 20\rm{cm}$ cloth, a $30\rm{cm} \times 30\rm{cm}$ cloth and a $35\rm{cm} \times 35\rm{cm}$ cloth. We vary the pose by placing the cloths in 3 rotation angles ($0^{\circ}$, $30^{\circ}$, and $45^{\circ}$). We also use a T-shirt ($50\rm{cm} \times 35\rm{cm}$) and a pair of shorts ($40\rm{cm} \times 32\rm{cm}$) which are unseen during training. All clothes can be found in Fig.~\ref{expermental setup}. We complete the same four tasks in simulation using the 3 square cloths. In each task, given a sequence of demonstration images where the cloth's size is $30\rm{cm} \times 30\rm{cm}$, and the rotation angle is $0^{\circ}$, Foldsformer should perform the task on all 9 combinations of different cloth size and pose. We also test the generalization of Foldsformer by completing a T-shirt folding task and a shorts folding task where the T-shirt and the shorts (see Fig.~\ref{real world figure}) are unseen cloth shapes during training. Similar to previous works~\cite{lee,fabricflownet,wrinkle}, we evaluate the performance quantitatively by using Mean Intersection over Union (MIoU) between the cloth masks achieved by Foldsformer and a human demonstrator and a penalty for wrinkles (WR) that calculates the fraction of pixels inside the cloth mask detected as edges by the Canny edge detector~\cite{canny}.

\begin{table}[b]
\centering
\caption{MIoU and mean WR on six sequential multi-step cloth manipulation tasks in the real world}
\setlength{\tabcolsep}{8mm}{\begin{tabular}{@{}lcc@{}}
\toprule
Task                 & MIoU & mean WR \\ \midrule
\textit{Double Triangle}      & 0.885 & 0.022 \\
\textit{Double Straight}      & 0.872 & 0.035\\
\textit{All Corners Inward}   & 0.911 & 0.042 \\
\textit{Corners Edges Inward} & 0.923 & 0.044\\ 
\textit{T-shirt Folding}      & 0.801 & 0.106\\
\textit{Shorts Folding}      & 0.676 & 0.061\\ \bottomrule
\end{tabular}}
\label{real world results}
\end{table}

\subsection{Real World Results}
Table.~\ref{real world results} shows the quantitative results, and Fig.~\ref{real world figure} shows part of our representative qualitative results. As can be seen, Foldsformer performs all the tasks successfully, even when the cloth’s size and pose differ from the cloth in demonstration sub-goals. Foldsformer can also generalize to the T-shirt and shorts, which are unseen during training. By accessing the complete picture of the task that contains all sub-goals together, Foldsformer can learn the instruction information and be robust to the variation of cloth configurations.
Our encoder enables the instruction information learned from a general demonstration can guide Foldsformer perform similar tasks on different cloth configurations. Hence, only a general demonstration is required by human demonstration and can be applied to different settings, which is more practical and efficient than previous approaches. Besides, since the cloth is often self-occluded and partially observable, providing all sub-goals together is helpful for an agent to obtain the global knowledge of a multi-step task and then infer proper chained actions. Videos of robot executions can be found on our project website.

\begin{figure}[b]
    \centering
    \includegraphics[width=0.45\textwidth]{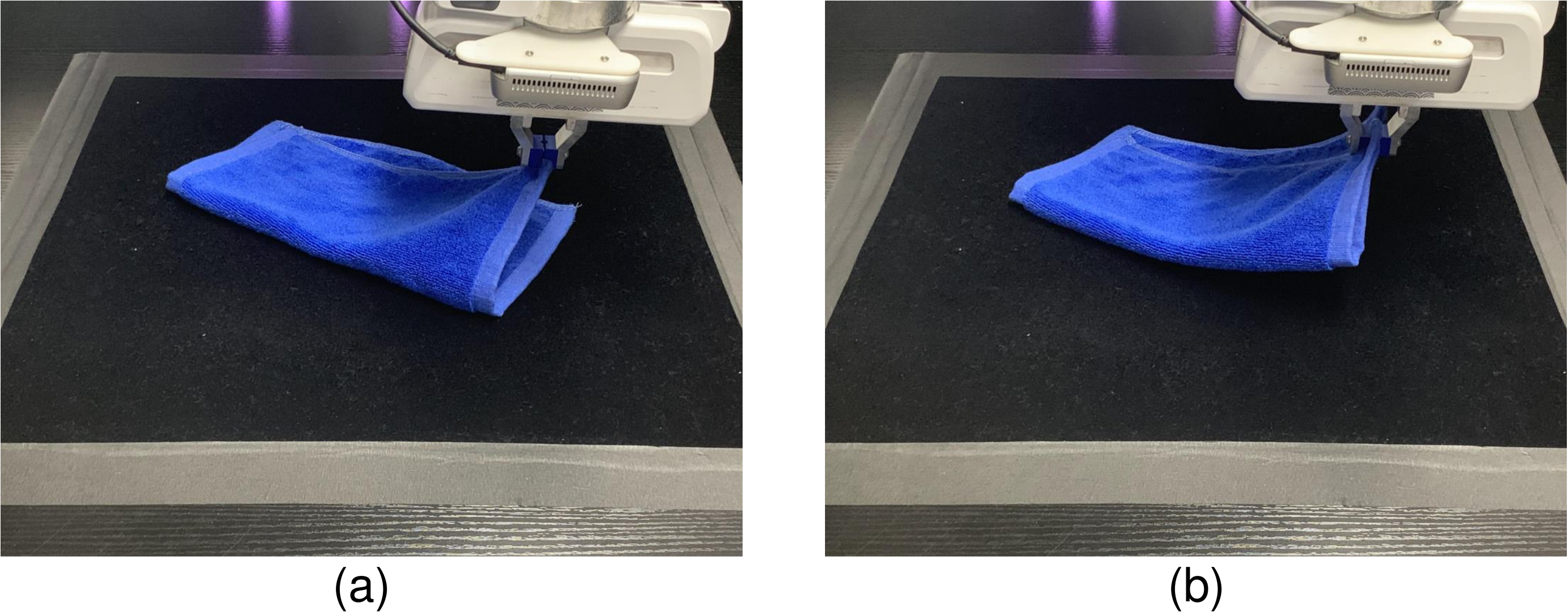}
    \setlength{\abovecaptionskip}{-0.1cm}
    \caption{\textbf{Failure Case.} Failure cases in our experiments mainly come from grasping failures. (a) Desired Grasp. The gripper is supposed to grasp one layer of the towel. (b) Executed Grasp. The gripper grasps two layers of the towel.}
    \label{failure case}
\end{figure}

\section{Limitations and Future Work}
Failure cases in our experiments mainly come from grasping failures. Since we use a pick-and-place action primitive in 2D pixel space, it's hard to determine the number of the grasped cloth layers. 
This grasp quality is crucial in the real-world setting, especially for \textit{Double Straight} and \textit{Corners Edges Inward} tasks. 
Fig.~\ref{failure case} illustrates this failure case. 
The cloth size should also be much smaller than the robot arm's workspace due to physical reachability limitations.
Meanwhile, the sim-to-real transfer relies on color-thresholding the background, which is not feasible in some cluttered backgrounds.
Further, although Foldsformer can successfully perform tasks on clothes of different configurations with a single general demonstration, the demonstration sub-goals still need to be generated by expert demonstrations.
In future work, we will try to make a fusion of visual information and tactile signals~\cite{tactile} to improve the grasp quality and reliability. We will also explore generating the sub-goals automatically.

\section{Conclusion}
In this paper, we present a novel framework named Foldsformer for sequential multi-step cloth manipulation. We utilize the space-time attention mechanism to capture the instruction information behind a general demonstration that can guide Foldsformer complete similar tasks on different cloth configurations.
Experiments results indicate that Foldsformer is effective and practical for sequential multi-step cloth manipulation tasks both in simulation and in the real world. Furthermore, Foldsformer can also generalize to unseen cloth shapes despite only being trained on rectangular cloths.

\bibliographystyle{IEEEtran}
\bibliography{citation} 

\end{document}